\documentclass{article}
\usepackage[utf8]{inputenc}
\DeclareUnicodeCharacter{221A}{$\sqrt{}$}
\usepackage{amsmath}
\usepackage{bm}
\usepackage{caption}
\usepackage[preprint]{neurips_2025}
\usepackage[table]{xcolor}
\usepackage{multirow}
\usepackage{graphicx}
\usepackage{tikz}
\usepackage{comment} 
\usepackage{makecell}

\usepackage[utf8]{inputenc} 
\usepackage[T1]{fontenc}    
\usepackage{hyperref}       
\usepackage{url}            
\usepackage{booktabs}       
\usepackage{amsfonts}       
\usepackage{nicefrac}       
\usepackage{microtype}      
\usepackage{xcolor}         

\DeclareUnicodeCharacter{FF1A}{:} 

\title{TACOcc:Target-Adaptive Cross-Modal Fusion with Volume Rendering for 3D Semantic Occupancy}

\author{%
  Luyao Lei\thanks{Corresponding author: Xing Wei.} \\
  School of Software Engineering\\
  Xi'an Jiaotong University\\
  Xi'an 710049, China \\
  \texttt{leiluyao@stu.xjtu.edu.cn} \\
  \And
  Shuo Xu \\
  School of Software Engineering\\
  Xi'an Jiaotong University\\
  Xi'an 710049, China \\
  \texttt{spoon1116@stu.xjtu.edu.cn} \\
  \And
  Yifan Bai \\
  School of Software Engineering\\
  Xi'an Jiaotong University\\
  Xi'an 710049, China \\
  \texttt{yfbai@stu.xjtu.edu.cn} \\
  \And
  Xing Wei\thanks{Corresponding author.} \\
  School of Software Engineering\\
  Xi'an Jiaotong University\\
  Xi'an 710049, China \\
  \texttt{weixing@mail.xjtu.edu.cn} \\
}

\begin{document}

\maketitle

\begin{abstract}
The performance of multi-modal 3D occupancy prediction is limited by ineffective fusion, mainly due to geometry-semantics mismatch from fixed fusion strategies and surface detail loss caused by sparse, noisy annotations. The mismatch stems from the heterogeneous scale and distribution of point cloud and image features, leading to biased matching under fixed neighborhood fusion. To address this, we propose a target-scale adaptive, bidirectional symmetric retrieval mechanism. It expands the neighborhood for large targets to enhance context awareness and shrinks it for small ones to improve efficiency and suppress noise, enabling accurate cross-modal feature alignment. This mechanism explicitly establishes spatial correspondences and improves fusion accuracy. For surface detail loss, sparse labels provide limited supervision, resulting in poor predictions for small objects. We introduce an improved volume rendering pipeline based on 3D Gaussian Splatting, which takes fused features as input to render images, applies photometric consistency supervision, and jointly optimizes 2D-3D consistency. This enhances surface detail reconstruction while suppressing noise propagation. In summary, we propose \textbf{TACOcc}, an adaptive multi-modal fusion framework for 3D semantic occupancy prediction, enhanced by volume rendering supervision. Experiments on the nuScenes and SemanticKITTI benchmarks validate its effectiveness.

\end{abstract}

\section{Introduction}

3D semantic occupancy prediction provides fine-grained geometric and semantic representations for real-time environmental perception in autonomous driving by densely parsing voxelized 3D spaces~\cite{cao2022monoscene,li2023voxformer,li2023fb,ramakrishnan2020occupancy,shepel2021occupancy,tan2025geocc,tong2023scene,wei2023surroundocc}. Compared to traditional 3D object detection, which is constrained by rigid bounding box representations, this technique effectively models irregular obstacles (such as collapsed vegetation or construction barriers) and continuous surface structures (such as road undulations). It demonstrates unique advantages in key tasks like dynamic obstacle avoidance and drivable area segmentation~\cite{wei2023surroundocc,xu2025survey,zhang2024vision}. However, current methods face two major bottlenecks in utilizing multi-modal data to enhance scene understanding: (1) the cross-modal fusion mechanism based on fixed-sampling neighborhoods leads to a spatial mismatch between geometric features and semantic representations~\cite{lei2023recent,ming2025inverse++,priyasad2021memory,zhang2024fusionocc}. (2) sparse annotated data causes cross-modal feature degradation, limiting the accuracy of reconstructing continuous surface details~\cite{chen2025tgp}.

To address the geometric-semantic mismatch problem, recent research attempts to dynamically adjust modality weights using attention mechanisms~\cite{huang2023tri}. However, these methods essentially perform global filtering along the feature channel dimension, applying the weight matrix to the entire feature map. They fail to capture local geometric-semantic correlation differences at different spatial locations, and thus do not fundamentally solve the spatial misalignment problem.

For the problem of missing details, emerging 3D Gaussian splatting (3DGS) techniques~\cite{fu2024colmap,huang2024gaussianformer,kerbl20233d} build photometric consistency constraints through differentiable volume rendering, theoretically enabling the compensation of sparse annotations using dense image signals. However, their single-modal architecture limits the exploration of complementary advantages in cross-modal features.

To solve the above limitations, we propose a new 3D semantic occupancy prediction framework named \textbf{TACOcc}, which uses a dual-module optimization mechanism for adaptive multi-modal fusion under volumetric rendering supervision. For the first problem, we design an adaptive fusion module that predicts optimal neighborhood ranges based on query features: the range is expanded for large objects to improve context awareness and reduced for small objects to suppress noise. This adaptive control guides a symmetric retrieval unit to perform two-way cross-modal interaction, allowing accurate alignment of multi-modal features. For the second problem, we introduce a volumetric rendering optimization module, which uses a 3DGS rendering pipeline with multi-modal inputs. This module calculates multi-view photometric reprojection loss from the input images to update the parameters of Gaussian primitives through optimization. We also add a Gaussian parameter consistency loss to build a learning path from 2D supervision signals to the 3D feature space. This method effectively prevents surface detail loss during fusion and improves the prediction of small objects in 3D occupancy tasks.

We conduct comprehensive experiments on the nuScenes and SemanticKITTI benchmarks to demonstrate the effectiveness of the proposed method. The TACOcc method achieves a 28.4\% mIoU in 3D semantic occupancy prediction, surpassing the current SOTA multi-modal method, Co-Occ, by 1.8\%. This confirms the effectiveness of the dual-module collaborative framework and provides a scalable new framework for multi-modal autonomous driving perception.

Our contributions are as follows: (1) We propose an adaptive feature fusion module that constructs a differentiable optimization strategy to dynamically generate the optimal neighborhood ranges. This drives a bidirectional, symmetric retrieval mechanism that adapts to the target scale, enabling efficient and precise alignment of multi-modal features in complex driving scenarios. (2) We introduce a volume rendering optimization module that replaces traditional hand-crafted initialization with multi-modal fusion initialization, and utilizes multi-view observation data to establish an optimized link from 2D image space to 3D occupancy space. (3) We propose an entirely new framework that achieves 28.4\% mIoU in the 3D semantic occupancy prediction task, setting a new SOTA.

\section{Related Work}

\textbf{3D Semantic Occupancy Prediction.}The core challenge lies in effectively representing spatial occupancy states and their semantic categories in 3D scenes. Early works focused on indoor scenarios ~\cite{liu2018see,song2017semantic,zhang2018efficient}, later shifting to outdoor environments using sparse lidar data ~\cite{cheng2021s3cnet,yan2021sparse,yang2021semantic}. MonoScene ~\cite{cao2022monoscene} pioneered monocular image-based outdoor occupancy prediction. Recent advances leverage multi-modal fusion for superior performance ~\cite{pan2024co,ramakrishnan2020occupancy,shepel2021occupancy,wang2023openoccupancy}, with OpenOccupancy ~\cite{wang2023openoccupancy} establishing the first benchmark for driving scenes. However, static-weight fusion mechanisms struggle with generalization in complex real-world environments. Adaptive multi-modal fusion approaches fall into three categories: (1) attention-based fusion ~\cite{liu2023multi,zhang2023provable}: dynamically weights modalities (e.g., prioritizing lidar when visual clarity degrades). (2) adaptive modality selection ~\cite{sahu2019dynamic,wang2022multi}: selects sub-networks based on input conditions (e.g., weather-specific processing branches). (3) gating mechanisms ~\cite{huang2023effective,mees2016choosing,xue2023dynamic}: regulates modality propagation paths during conflicts (e.g., suppressing erroneous modalities). Our work aligns with category 1 and the Co-Occ method ~\cite{pan2024co} relevant to this study has significant limitations. It employs a fixed neighborhood search range. This approach fails to fully align the geometric features of point clouds with the semantic features of images. Moreover, to meet the perception needs of large-sized targets, it has to use a larger search range. This inevitably introduces many irrelevant voxels for computation in small target detection scenarios. So we introduce a target-adaptive bidirectional retrieval mechanism via joint optimization of Straight-Through Estimator (STE) and Gumbel-Softmax, resolving feature misalignment through precise cross-modal matching.

\textbf{Volume Rendering for Scene Understanding.}Volume rendering enhances geometric fidelity by optimizing continuous volumetric representations under photometric supervision. Traditional methods ~\cite{kundu2022panoptic,wu2023mars} lack flexibility for cross-scene generalization. ~\cite{gan2023simple,wimbauer2023behind} discussed rendering-based occupancy prediction but did not consider semantic information. Later methods, such as ~\cite{pan2024renderocc,pan2023uniocc}, used Neural Radiance Fields (NeRF) ~\cite{mildenhall2021nerf} to predict 3D semantic occupancy, achieving improved anti-aliasing effects~\cite{barron2021mip}, refined reflectance~\cite{verbin2022ref}, training from sparse views ~\cite{irshad2023neo}, fewer training iterations~\cite{muller2022instant}, and reduced rendering time~\cite{garbin2021fastnerf}. However, rendering speed still limits their prediction efficiency. Recent advances favor point-based representations ~\cite{kerbl20233d,zhang2022differentiable} for efficiency. ~\cite{zhang2022differentiable} employs a differentiable splat-based renderer for cross-view point optimization. 3DGS~\cite{kerbl20233d} revolutionizes real-time novel view synthesis through explicit voxel modeling and GPU-optimized differentiable splatting. However, traditional 3DGS suffers from two critical limitations: (1) Single-Modal Dependency: reliance on single-modal inputs constrains geometric priors. (2) Empirical Initialization: manual initialization based on point cloud density significantly increases subsequent optimization costs. Inspired by 3DGS, our work uses an improved 3DGS that supports multi-modal information processing and auto-initialization in the volumetric rendering optimization module. It effectively solves the problems of the loss of feature details and noise interference after fusion, and improves the occupancy prediction accuracy. 

\section{Method}
We use lidar point clouds and their surround-view images as inputs to predict 3D semantic occupancy in driving scenarios. Our framework consists of two key components: the adaptive feature fusion module and the volume rendering optimization module. As shown in Fig.\ref{fig:figure1}:

\begin{figure}[h]
\includegraphics[
    width=1\textwidth,
    trim={0 60 0 30}, 
    clip
]{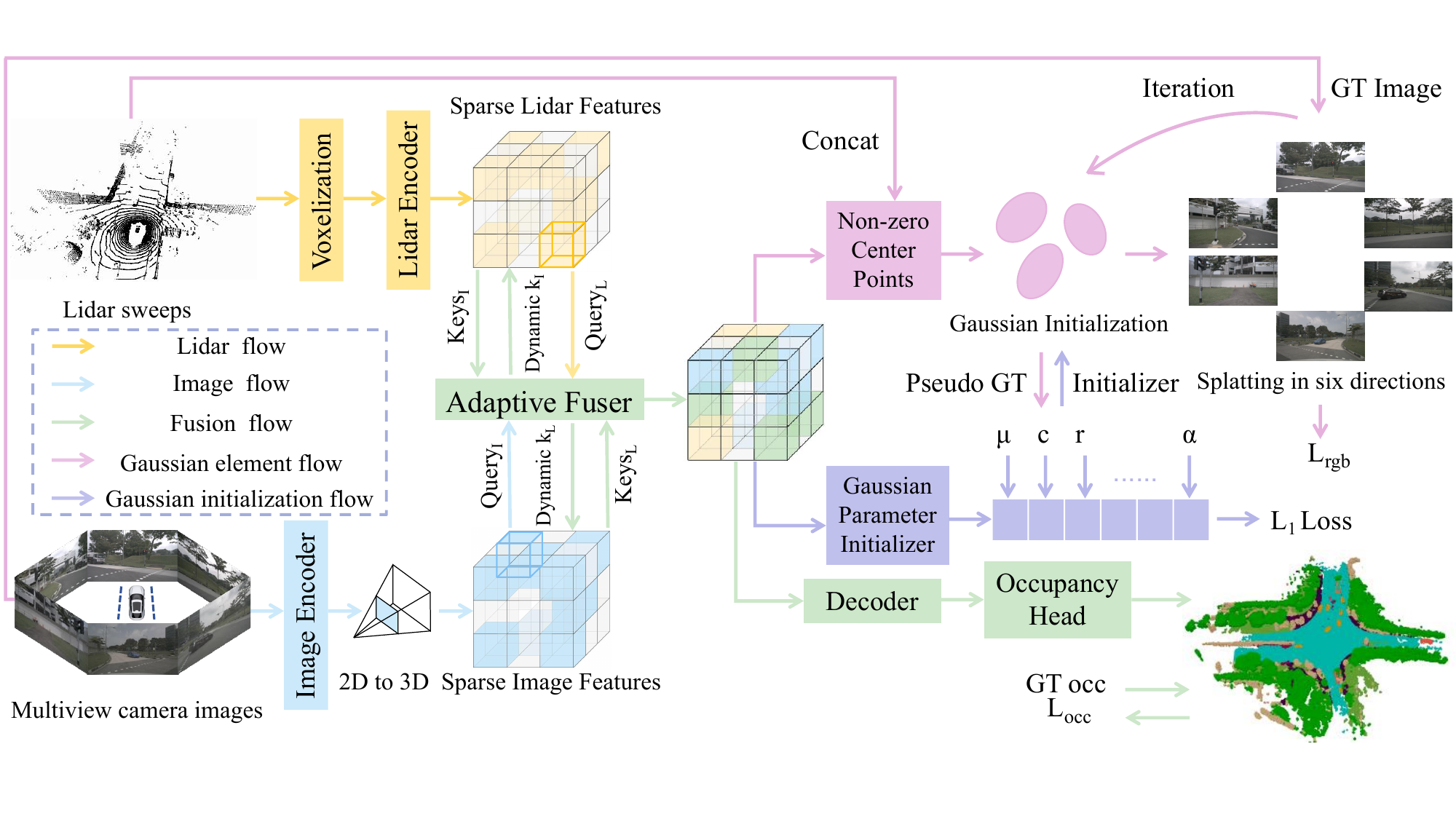}
\centering
\caption{Overview of proposed TACOcc. The point clouds and images undergo feature extraction and dimensional transformation to generate sparse voxel features. These features are then precisely aligned and fused through an adaptive fusion module (See Sec.\ref{sec:3.1}), followed by enhancement via a volume rendering optimization module (See Sec.\ref{sec:3.2}), thereby improving the performance of 3D semantic occupancy prediction.}
\label{fig:figure1}
\end{figure}

This framework first generates sky region masks using an unsupervised semantic segmentation algorithm to suppress non-informative areas in images, while applying conditional filtering to denoise raw lidar point clouds. The denoised point cloud data is voxelized, and a sparse convolutional neural network extracts voxel features. Simultaneously, the masked images are fed into an image encoder for feature extraction, with camera parameters and depth information enabling 2D-3D view transformation to project image features into the same 3D space as the point clouds. Then, we use the scale-adaptive fusion module that predicts adaptive neighborhood ranges based on non-zero query features. These ranges guide a symmetric retrieval unit to enable bidirectional cross-modal interaction. The module dynamically adjusts aggregation weights between image and point cloud features, achieving complementary enhancement of geometric and semantic information. We then inject the fused features into the volume rendering optimization module, establishing an optimization pathway from 2D supervision signals to the 3D feature space. Finally, the enhanced fused features are input into a cascaded occupancy head\cite{wang2023openoccupancy} to output a fine-grained 3D semantic occupancy grid.

\subsection{Adaptive Feature Fusion Module}
\label{sec:3.1}
The adaptive fusion module achieves more precise alignment of multi-modal features in complex driving scenarios through a target-adaptive bidirectional symmetric retrieval mechanism. The process is shown in Fig.\ref{fig:figure2}:

\begin{figure}[h]
\includegraphics[
    width=1\textwidth,
    trim={30 30 15 30}, 
    clip
]{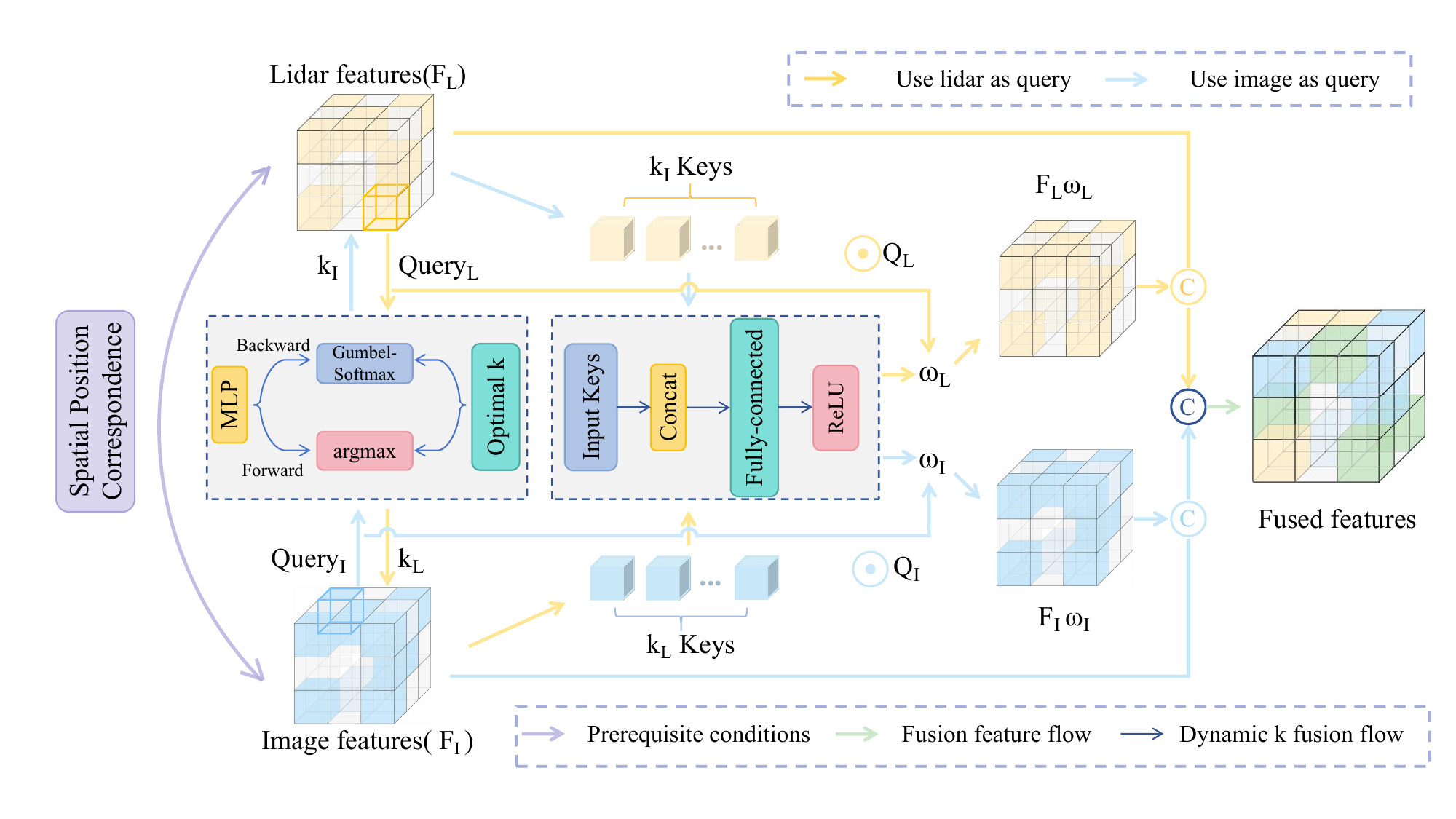}
\centering
\caption{Adaptive Fuser. To achieve target-aligned fusion of sparse voxel features from two modalities, a dynamic $k$ selection module generates an optimal $k$ value for each query. Based on this value, bidirectional symmetric retrieval is performed. The retrieved keys are stacked and passed through feature extraction and non-linear transformation to compute attention weights. These weights are then used to weight the current query, resulting in the fused feature.
}
\label{fig:figure2}
\end{figure}

This module constructs a nonlinear mapping from current query features to candidate integers $k$ using a multilayer perceptron (MLP). The core challenge in optimal $k$ selection stems from the parametric modeling of discrete probability distributions: traditional discrete sampling operators like argmax break gradient backpropagation due to their non-differentiable nature. To overcome this, we introduce the Gumbel-Softmax ~\cite{jang2016categorical}  reparameterization technique to establish a continuous relaxation mechanism. Gumbel-Softmax makes the discrete sampling process continuous by combining the Gumbel distribution and the Softmax function. Subsequently, the generated probability distribution is multiplied by the candidate $k$ to calculate the expectation. The expectation is then discretized via rounding to determine the optimal $k$ for the current scenario.

However, the rounding discretization operation can disrupt the integrity of the computational graph, causing a break in the gradient backpropagation chain at critical decision-making nodes. To address this, we employ the Straight-Through Estimator (STE)~\cite{yin2019understanding} to construct a surrogate gradient function(Formula \ref{eq:formula5}\ref{eq:formula4}). It decouples gradient computation between forward/backward propagations. Specifically, the discretized values after rounding are used during forward propagation, while continuous values are used during backward propagation to balance discreteness and differentiability.
\begin{equation}
\nabla_z \mathcal{L} \approx \nabla_{z_c} \mathcal{L}
\label{eq:formula5}
\end{equation}
\begin{equation}
\frac{\partial \mathcal{L}}{\partial z} = 
\begin{cases} 
I, & \text{Forward: } y_{\text{ste}} = e_k\ (k = \arg\max(z)) \\
\frac{\partial z_c}{\partial z}, & \text{Backward: } z_c = \mathrm{GumbelSoftmax}(z)
\end{cases}
\label{eq:formula4}
\end{equation}

where $z$ denotes the logits produced by the MLP, \(\nabla_{z}\mathcal{L}\) is the gradient of the loss function with respect to the discrete variable \(z\), \(\nabla_{z_{c}}\mathcal{L}\) is the gradient of the loss function with respect to the continuous relaxed variable \(z_{c}\), \(I\) is the identity matrix. During the forward propagation, $y_{\text{ste}}$ is computed based on the discrete $z$, where the class index is determined via $k=\text{argmax}(z)$, generating the corresponding one-hot encoded vector $e_k$. In backward propagation, to enable gradient updates, we use the continuous Gumbel-Softmax output $z_c = \text{GumbelSoftmax}(z)$ as a differentiable approximation for the discrete $z$. By combining the STE with Gumbel-Softmax, the approach preserves gradient propagation while effectively addressing the discretization problem of selecting the optimal $k$ via argmax. 

Based on the current optimal $k$, the system implements an adaptive neighborhood sampling strategy for each valid query voxel in the cross-modal aligned feature space. Using the K-Nearest Neighbors method, captures top-$k$ non-zero voxels from the target modality as key vectors. Specifically, for the i-th non-zero query feature, its adjacent non-zero features are represented as $\{N_i^0, N_i^1, \ldots, N_i^{k_i}\}$, where $k_i$ is the $k$ corresponding to the current query. It stacks the $k_i$ feature vectors to form key matrix $K_i$, which contains important information related to the query feature and provides a basis for subsequent feature enhancement and fusion. Then, a fully connected layer is used to perform a linear mapping of the feature space of the key matrix $K_i$, and a nonlinear activation function is combined to generate selective attention weights (formula \ref{eq:formula7}). This weight matrix enhances the features by weighting the original query features. Meanwhile, a symmetric architecture is adopted at the other modality end to implement reverse feature enhancement. This bidirectional interaction paradigm significantly improves the robustness and semantic discriminative power of cross-modal representations.

\begin{equation}
\omega_i = \text{Linear}(\text{Padding}(\text{Concat}(N_i^0, N_i^1, \dots,  N_i^{k_i})))
\label{eq:formula7}
\end{equation}

Finally, the weighted query features are concatenated with the original image features and point cloud features (Formula \ref{formula44}). This concatenation operation integrates feature information from different sources, providing a richer feature representation for subsequent processing steps.

\begin{equation}
F_{IL} = \text{Concat}(F_I, F_L, F_I \cdot \omega_I, F_L \cdot \omega_L)
\label{formula44}
\end{equation}

\subsection{Volume Rendering Optimization Module}
\label{sec:3.2}
We use volume rendering technology to solve the problem of missing continuous surface details and noise interference in fusion caused by sparse annotations, and further optimize the fused features. We propose a multi-modal-enhanced 3DGS architecture with two key innovations: 
(1) Multi-Source Feature-Driven Parameter Initialization: a lightweight 3D convolutional network maps fused features into Gaussian primitives parameters. (2) Cross-Modal Geometric Anchor Coupling: beyond raw point cloud coordinates, we also introduce non-zero voxel centers filtered by fused features as auxiliary anchor points to enhance scene coverage completeness.

During the parameter initialization stage, we utilize the fused multi-source features to obtain the initial parameters of each Gaussian component through a lightweight convolutional network. Subsequently, the system constructs a Gaussian field using two inputs: the point cloud coordinates filtered by the voxel grid and the non-zero voxel centers derived from the fused features. Specifically, point cloud coordinates offer high-precision 3D geometric information but may exhibit sparsity in certain regions, resulting in the loss of local details. In contrast, voxel coordinates divide the space into regular grids, thereby compensating for the missing regional information in the point cloud and improving the overall coverage and completeness of the scene. This complementary relationship helps prevent coverage blind spots that could arise from the limitations of relying on a single modality.
We employ differentiable rendering techniques \cite{kerbl20233d} to project 3D Gaussians onto the 2D image space of six viewpoints. Then, we use a neural point-based method \cite{kopanas2022neural} to compute the color of each pixel by blending ordered points that overlap with each pixel. The rendered images are optimized with supervision from the input images to refine the Gaussian primitives parameters. Specifically, the optimization process is driven by a progressive diffusion mechanism, guiding the Gaussian model to adaptively grow along the geometric gradient direction. The model's parameters—such as color and scale—are iteratively optimized starting from an initial value, with a multi-view photometric consistency loss $\mathcal{L}_{rgb}$ guiding the optimization process. This dynamic adjustment modifies the radiative properties and geometric shape of each Gaussian primitive. Once the Gaussian parameters converge to a stable state, they are treated as pseudo-ground truth and compared with the initial parameters to compute the parameter consistency loss $\mathcal{L}_{pc}$. This step compensates for details that may be missing from the 2D input images in the fused 3D multi-modal features. 

During the rendering process, the multi-modal initialization strategy uses purely geometric coordinates as input. This avoids optimization ambiguities caused by feature coupling. Moreover, it solves coverage blind spots in complex scenes by leveraging the complementarity between point clouds and voxel coordinates.

\subsection{Optimization}
\textbf{Occupancy Loss.}
Use cross-entropy loss ($\mathcal{L}_{\text{ce}}$) and Lovász-softmax loss to supervise the relationship between predicted semantic occupancy and GT (ground truth) occupancy (Formula \ref{loss1})~\cite{pan2024co}.

\begin{equation}
\mathcal{L}_{\text{occ}} = \mathcal{L}_{\text{ce}} + \mathcal{L}_{\text{lovasz}}
\label{loss1}
\end{equation}

\textbf{Gaussian Parameter Consistency Loss.}
Calculated from the initial Gaussian parameters and the iteratively optimized Gaussian parameters. (Formula \ref{loss2})
\vspace{-0.8em}

\begin{equation}
\mathcal{L}_{pc} = \sum_{i=1}^{N} \left\| \Theta_{i}^{\text{final}} - \Theta_{i}^{(0)} \right\|_{1}
\label{loss2}
\end{equation}

\textbf{Volume Rendering Photometric Loss. }
A combination of L1 and D-SSIM used to supervise the relationship between rendered images and current real input images (Formula \ref{loss3})~\cite{fu2024colmap}.

\vspace{-0.8em}

\begin{equation}
\mathcal{L}_{\text{rgb}} = (1-\lambda)\mathcal{L}_1 + \lambda \mathcal{L}_{D-SSIM}
\label{loss3}
\end{equation}
\textbf{Total Loss.} During Gaussian parameter iteration, $\mathcal{L}_{\text{rgb}}$ updates only Gaussians; post-iteration, weighted $\mathcal{L}_{\text{occ}}$ and $\mathcal{L}_{\text{pc}}$ refine fused features for enhanced detail. (Formula \ref{loss4})
\begin{equation}
\mathcal{L}_{\mathrm{total}} = 
\begin{cases}
\mathcal{L}_{\mathrm{rgb}} & \text{During Iteration} \\
(1-\lambda)\mathcal{L}_{\mathrm{occ}} + \lambda\mathcal{L}_{\mathrm{pc}} & \text{After Iteration}
\end{cases}
\label{loss4}
\end{equation}

\section{Experiments}
\subsection{Dataset}

\textbf{NuScenes}~\cite{caesar2020nuscenes}. As a benchmark dataset for autonomous driving, its multi-modal spatio-temporal alignment and dynamic scene coverage capabilities support voxelized occupancy grid modeling. This dataset contains 1000 driving scenarios (850 for training / 150 for validation). The three-dimensional perception standard is established through the strict synchronization of a lidar, six cameras, and millimeter-wave radars. Compared with KITTI~\cite{geiger2013vision}, its advantages are as follows: 1) Dense 3D dynamic annotations covering 23 types of targets. 2) 20-second continuous trajectory data and radar motion compensation to solve dynamic deformation interference. 3) A multi-modal spatio-temporal alignment mechanism to support cross-modal geometric feature fusion. 4) Built-in sensor pose matrices and IMU data to provide spatio-temporal continuity priors. These characteristics make it an ideal platform for validating dynamic 3D occupancy prediction and optimal for joint inference tasks in complex urban scenarios.

\textbf{SurroundOcc}~\cite{wei2023surroundocc}. A 3D occupancy prediction dataset built based on nuScenes, which generates dense occupancy labels through multi-modal fusion and automated annotation. Its labels are generated by multi-frame point cloud stitching, Poisson reconstruction voxelization (with a resolution of 0.5m, covering a 200 × 200 × 16 grid), and semantic transfer, and it has the ability to model dynamic scenes. The core advantages include: 1) Full compatibility with nuScenes sensor parameters to avoid spatio-temporal misalignment. 2) An automated process to reduce annotation costs. Compared with Occ3D-nuScenes ~\cite{tian2023occ3d} and SemanticKITTI ~\cite{behley2019semantickitti}, it is superior in annotation density and dynamic modeling. It is suitable for validating fusion methods, evaluating the robustness of small target prediction, and providing a training benchmark for real-time models, significantly improving the generalization ability in the open world. 

\subsection{Implementation Details}
We evaluated the 3D semantic occupancy performance on the nuScenes validation set~\cite{caesar2020nuscenes}. Using a ResNet101~\cite{he2016deep} backbone and a 2D-to-3D view transformer~\cite{philion2020lift} to generate a 3D feature volume of size \(128 \times 128 \times 16\). The 10 lidar point clouds were voxelized using a voxel encoder. The adaptive fusion module fused the features with candidate \(k\) values of 1, 2, 3, 4. The same occupancy decoder and head as in ~\cite{wang2023openoccupancy} were used with a cascade ratio of 2 for refined predictions. The model was trained for 15 epochs on nuScenes using 8 RTX A6000 GPUs with a batch size of 8, completing training in 2 days. The AdamW~\cite{loshchilov2017decoupled} optimizer with a weight decay of 0.01 and an initial learning rate of \(1 \times 10^{-4}\), using a stepwise cosine decay, was employed. The D-SSIM window size was \(11 \times 11\), and \(\lambda\) was set to 0.2 in all experiments.

\subsection{Experimental Results}
We conduct experiments on the nuScenes~\cite{caesar2020nuscenes} dataset and compared our approach with several SOTA methods across different modalities (Tab.\ref{tab:result}), including camera-only methods~\cite{cao2022monoscene,li2203bevformer,pan2024renderocc,wang2023openoccupancy,wei2023surroundocc,zhang2023occformer}, lidar-only methods~\cite{roldao2020lmscnet,wang2023openoccupancy}, and lidar-camera fusion methods~\cite{pan2024co,wang2023openoccupancy}. Specifically, MonoScene~\cite{cao2022monoscene}, SurroundOcc~\cite{wei2023surroundocc}, and BEVFormer~\cite{li2203bevformer} used an input size of 900×1600 and a ResNet101-DCN backbone. In contrast, OccFormer~\cite{zhang2023occformer}, C-CONet~\cite{wang2023openoccupancy}, M-CONet~\cite{wang2023openoccupancy}, and RenderOcc~\cite{pan2024renderocc} used an input image size of 896×1600 with a 2D ResNet101 backbone. 
\begin{table}[h]
\centering
\vspace{-1.1em}
\caption{3D semantic occupancy prediction results on nuScenes validation.}
\vspace{0.5em}
\label{tab:result}
\resizebox{1.00\textwidth}{!}{%
{
\setlength{\tabcolsep}{3pt}
\begin{tabular}{ccccccccccccccccccccccc}
\toprule
\multirow{2}{*}{Method} & \multirow{2}{*}{Modality} & \multirow{2}{*}{IoU} & \multirow{2}{*}{mIoU} & \rotatebox{90}{barrier} & \rotatebox{90}{bicycle} & \rotatebox{90}{bus} & \rotatebox{90}{car} & \rotatebox{90}{const. veh.} & \rotatebox{90}{motorcycle} & \rotatebox{90}{pedestrian} & \rotatebox{90}{traffic cone} & \rotatebox{90}{trailer} & \rotatebox{90}{truck} & \rotatebox{90}{drive. suf.} & \rotatebox{90}{other flat} & \rotatebox{90}{sidewalk} & \rotatebox{90}{terrain} & \rotatebox{90}{manmade} & \rotatebox{90}{vegetation}& \multirow{2}{*}{Input Size} & \multirow{2}{*}{2D Backbone}  \\
 & &  &  & \textcolor[RGB]{255,120,50}{\rule{6pt}{6pt}} & \textcolor[RGB]{255,192,203}{\rule{6pt}{6pt}} & \textcolor[RGB]{255,255,0}{\rule{6pt}{6pt}} & \textcolor[RGB]{0,150,245}{\rule{6pt}{6pt}} & \textcolor[RGB]{0,255,255}{\rule{6pt}{6pt}} & \textcolor[RGB]{200,180,0}{\rule{6pt}{6pt}} & \textcolor[RGB]{255,0,0}{\rule{6pt}{6pt}} & \textcolor[RGB]{255,240,150}{\rule{6pt}{6pt}} & \textcolor[RGB]{135,60,0}{\rule{6pt}{6pt}} & \textcolor[RGB]{160,32,240}{\rule{6pt}{6pt}} & \textcolor[RGB]{255,0,255}{\rule{6pt}{6pt}} & \textcolor[RGB]{139,137,137}{\rule{6pt}{6pt}} & \textcolor[RGB]{75,0,75}{\rule{6pt}{6pt}} & \textcolor[RGB]{150,240,80}{\rule{6pt}{6pt}} & \textcolor[RGB]{230,230,250}{\rule{6pt}{6pt}} & \textcolor[RGB]{0,175,0}{\rule{6pt}{6pt}} &  &  \\
 \midrule
MonoScene~\cite{cao2022monoscene} & C & 22.4 & 7.2 & 6.4 & 2.2 & 8.6 & 7.1 & 4.5 & 2.3 & 2.8 & 1.9 & 4.0 & 4.4 & 16.3 & 6.2 & 12.5 & 10.3 & 9.7& 15.8 & 900×1600 & R101-DCN \\
SurroundOcc~\cite{wei2023surroundocc} & C & 31.5 & 20.1 & 19.3 & 11.8 & 29.1 & 10.0 & 15.4 & 13.9 & 12.3 & 14.1 & 21.3 & 37.3 & 24.7 & 24.2 & 17.4 & 23.3 & 23.9 & 24.9 & 900×1600 & R101-DCN \\
BEVFormer~\cite{li2203bevformer} & C & 30.7 & 16.8 & 13.9 & 6.3 & 23.8 & 27.9 & 8.3 & 10.4 & 6.9 & 4.8 & 10.8 & 18.2 & 38.1 & 19.2 & 22.7 & 21.6 & 14.1 & 22.0 & 900×1600 & R101-DCN \\
C-CONet~\cite{wang2023openoccupancy} & C & 25.7 & 18.6 & 18.7 & 10.3 & 27.8 & 27.7 & 8.2 & 16.3 & 13.8 & 9.6 & 11.3 & 19.2 & 33.2 & 20.5 & 22.1 & 21.6 & 14.9 & 22.6 & 896×1600 & R101 \\
OccFormer~\cite{zhang2023occformer} & C & 30.1 & 19.9 & 29.8 & 11.5 & 28.4 & 30.8 & 10.3 & 16.3 & 13.9 & 12.6 & 14.6 & 20.7 & 36.8 & 22.3 & 23.5 & 22.1 & 14.2 & 20.6 & 896×1600 & R101 \\
RenderOcc~\cite{pan2024renderocc} & C & 29.0 & 18.7 & 18.9 & 11.3 & 27.9 & 9.9 & 13.7 & 13.8 & 11.9 & 12.7 & 20.5 & 32.5 & 21.2 & 24.0 & 21.5 & 22.1 & 15.2 & 21.6 & 896×1600 & R101 \\
LMSCNet~\cite{roldao2020lmscnet} & L & 32.6 & 13.2 & 12.8 & 4.2 & 13.4 & 19.1 & 10.6 & 4.9 & 7.3 & 7.5 & 10.4 & 11.3 & 23.3 & 12.7 & 16.5 & 13.9 & 15.1 & 28.0 &-&-\\
L-CoNet~\cite{wang2023openoccupancy} & L & 39.1 & 18.1 & 19.0 & 3.9 & 16.2 & 27.3 & 7.3 & 4.1 & 7.2 & 5.8 & 14.6 & 14.3 & 38.9 & 20.2 & 24.5 & 24.2 & \textbf{25.1} & 36.9 &-&-\\
M-CoNet~\cite{wang2023openoccupancy} & C\&L & 39.2 & 23.6 & 23.4 & 12.4 & 31.1 & 33.2 & 14.1 & 17.2 & 18.7 & 13.4 & 19.9 & 25.2 & 38.2 & 21.2 & 25.4 & 25.1 & 24.7 & 34.9 & 896×1600 & R101  \\
LC-Fusion~\cite{ma2024licrocc} & C\&L & 40.7 & 25.0 & 27.8 & 16.3 & \textbf{33.7} & 33.2 & 17.5 & 17.2 & 18.7 & 15.4 & 19.5 & 25.2 & 39.1 & 21.2 & 22.3 & \textbf{29.1} & \textbf{29.6} & 34.9 & 896×1600 & R101  \\
Co-Occ~\cite{pan2024co} & C\&L & 41.0 & 26.6 & 27.3 & 16.5 & 33.5 & 37.5 & 17.9 & 21.8 & 16.8 & 21.7 & \textbf{27.6} & 39.2 & 25.9 & \textbf{28.1} & 29.2 & 26.2 & 18.9 & 36.8 & 896×1600 & R101  \\

\midrule
TACOcc(Ours) & C\&L & \textbf{41.8} & \textbf{28.4} & \textbf{30.1} & \textbf{18.5} & 31.2 & \textbf{38.6} & \textbf{18.6} & \textbf{23.1} & \textbf{20.9} & \textbf{23.8} & 25.6 & \textbf{39.3} & \textbf{40.3} & 26.7 & \textbf{29.8} & 26.6 & 24.1 & \textbf{37.6} & 896×1600 & R101  \\
\bottomrule
\end{tabular}%
}
}

\end{table}
\vspace{-0.66em}
\begin{figure}[h]
\includegraphics[
    width=1\textwidth,
    trim={140pt 0 160pt 0}, 
    clip
]{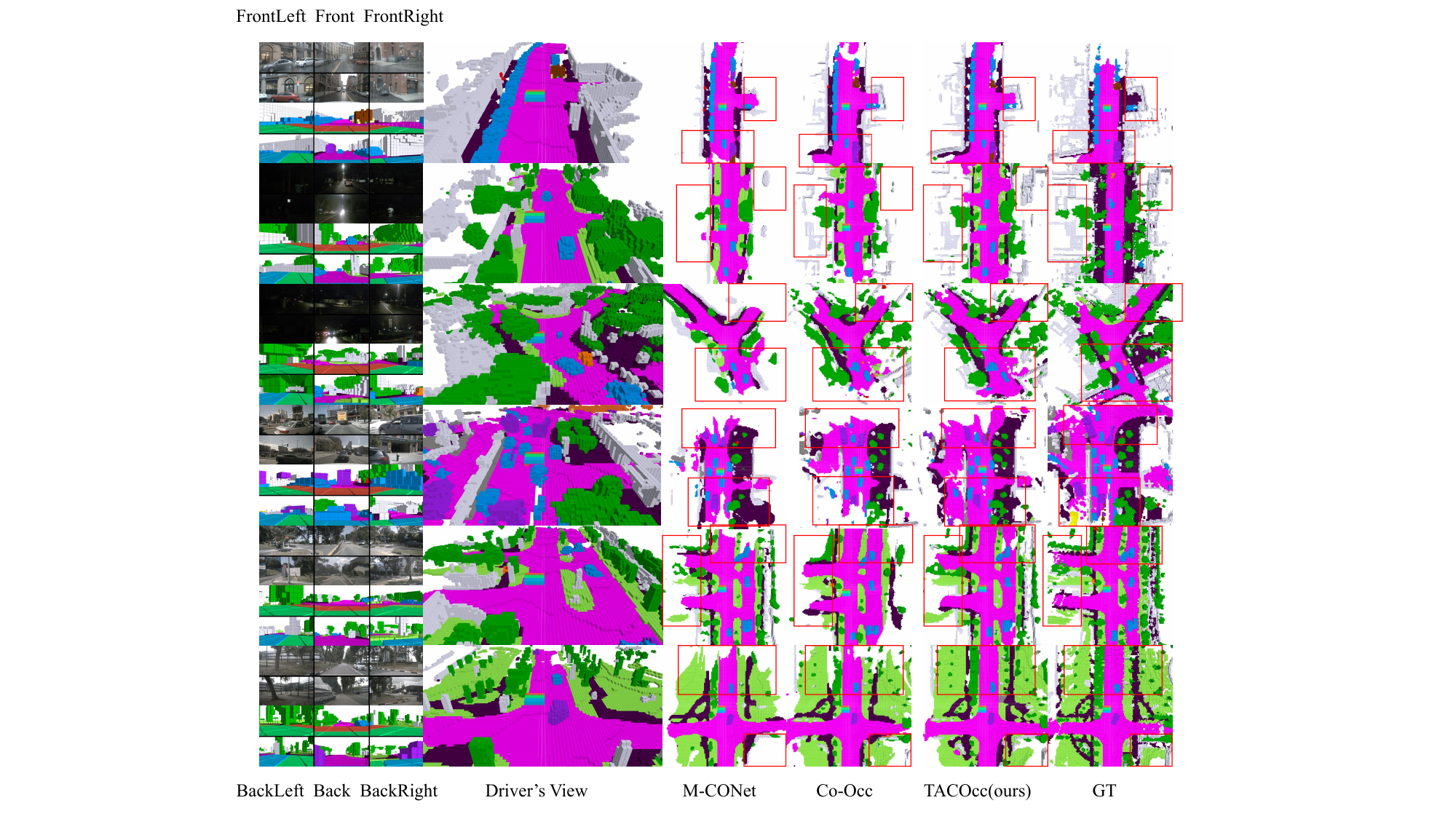}
\centering
\begin{scriptsize}
\textcolor[RGB]{255,0,255}{\rule{5pt}{5pt}} driveable surface \quad
\textcolor[RGB]{0,150,245}{\rule{5pt}{5pt}} car \quad
\textcolor[RGB]{255,255,0}{\rule{5pt}{5pt}} bus  \quad
\textcolor[RGB]{160,32,240}{\rule{5pt}{5pt}} truck \quad
\textcolor[RGB]{150,240,80}{\rule{5pt}{5pt}} terrain \quad
\textcolor[RGB]{0,175,0}{\rule{5pt}{5pt}} vegetation \quad
\textcolor[RGB]{75,0,75}{\rule{5pt}{5pt}} sidewalk \quad
\textcolor[RGB]{139,137,137}{\rule{5pt}{5pt}} other flat \quad
\textcolor[RGB]{255,0,0}{\rule{5pt}{5pt}} pedestrian \quad
\textcolor[RGB]{255,192,203}{\rule{5pt}{5pt}} bicycle \quad
\textcolor[RGB]{230,230,250}{\rule{5pt}{5pt}} manmade \quad
\textcolor[RGB]{200,180,0}{\rule{5pt}{5pt}} motorcycle \quad
\textcolor[RGB]{255,120,50}{\rule{5pt}{5pt}} barrier \quad
\textcolor[RGB]{0,255,255}{\rule{5pt}{5pt}} construction vehicle \quad
\textcolor[RGB]{135,60,0}{\rule{5pt}{5pt}} trailer \quad
\textcolor[RGB]{255,240,150}{\rule{5pt}{5pt}} traffic cone \quad
\vbox{%
  \offinterlineskip
  \hbox{\color[RGB]{227,83,60}\vrule width 5pt height 1.66666pt depth 0pt}
  \hbox{\color[RGB]{0,220,110}\vrule width 5pt height 1.66667pt depth 0pt}
  \hbox{\color[RGB]{0,166,221}\vrule width 5pt height 1.66667pt depth 0pt}
} ego vehicle
\end{scriptsize}
\caption{Qualitative comparison results on the nuScenes validation set. The top left displays the input surrounding images, while the bottom left shows the occupancy predictions made by our TACOcc for each surrounding image. Subsequently visualized are the driver's perspective from our TACOcc method, the Bird-Eye View (BEV) of M-CONet~\cite{wang2023openoccupancy}, the BEV of Co-Occ~\cite{pan2024co}, the BEV of our TACOcc method, and the ground truth (GT) occupancy from SurroundOcc~\cite{wei2023surroundocc}.}
\label{fig:figure3}
\vspace{-0.5cm}
\end{figure}

Our results show that the proposed TACOcc method achieves a 3D semantic occupancy prediction mIoU of 28.4\%, outperforming the previous best multi-modal method, Co-Occ~\cite{pan2024co}, by 1.8\%. It improves geometric-semantic consistency for small objects such as pedestrians and cyclists. Specifically, it improves the IoU of sparse point cloud objects such as traffic cones and motorcycles by nearly 3\%, while maintaining a high IoU for medium and large vehicles. At the same time, it effectively predicts distant and small objects.

We present the visualization results on nuScenes dataset in Fig.\ref{fig:figure3}. These results demonstrate the advantages of our method over other multi-modal approaches: (1) With the adaptive fusion module, our model dynamically adjusts the weights of image and point cloud features based on the characteristics of the input data. This enables significantly improved prediction performance under challenging scenarios such as rainy weather (where lidar data is susceptible to noise) and nighttime conditions (where image quality degrades). (2) With the volume rendering optimization module, our model establishes an optimized 2D-3D pathway by integrating fused features with the input images. This enhances the detail representation in the fused features, thereby improving 3D occupancy prediction performance, particularly for small objects.

\begin{table}[h]
\centering
\vspace{1.0em}
\vspace{-1em}
\caption{3D semantic occupancy prediction results on SematicKITTI test set.}
\label{tab:result2}
\vspace{0.5em}
\resizebox{1.0\textwidth}{!}{
{
\setlength{\tabcolsep}{3pt}
\begin{tabular}{ccccccccccccccccccccccc}
\toprule
\multirow{2}{*}{Method} & \multirow{2}{*}{Modality} & \multirow{2}{*}{mIoU} 
& \rotatebox{90}{\makecell[l]{\textbf{road} \\ \scriptsize{(13.90\%)}}}  & \rotatebox{90}{\makecell[l]{\textbf{sidewalk}\\ \scriptsize{(11.3\%)}}} & \rotatebox{90}{\makecell[l]{\textbf{parking} \\ \scriptsize{(1.12\%)}}} 
& \rotatebox{90}{\makecell[l]{\textbf{other ground} \\ \scriptsize{(0.56\%)}}} & \rotatebox{90}{\makecell[l]{\textbf{building} \\ \scriptsize{(14.1\%)}}} & \rotatebox{90}{\makecell[l]{\textbf{car} \\ \scriptsize{(9.92\%)}}} 
& \rotatebox{90}{\makecell[l]{\textbf{truck} \\ \scriptsize{(0.64\%)}}} & \rotatebox{90}{\makecell[l]{\textbf{bicycle} \\ \scriptsize{(0.03\%)}}} & \rotatebox{90}{\makecell[l]{\textbf{motorcycle} \\ \scriptsize{(0.03\%)}}} 
& \rotatebox{90}{\makecell[l]{\textbf{other vehicle} \\ \scriptsize{(0.20\%)}}} & \rotatebox{90}{\makecell[l]{\textbf{vegetation} \\ \scriptsize{(39.3\%)}}} & \rotatebox{90}{\makecell[l]{\textbf{trunk} \\ \scriptsize{(0.51\%)}}} 
& \rotatebox{90}{\makecell[l]{\textbf{terrain} \\ \scriptsize{(1.79\%)}}} & \rotatebox{90}{\makecell[l]{\textbf{person} \\ \scriptsize{(0.07\%)}}} & \rotatebox{90}{\makecell[l]{\textbf{bicyclist} \\ \scriptsize{(0.07\%)}}} 
& \rotatebox{90}{\makecell[l]{\textbf{motorcyclist} \\ \scriptsize{(0.03\%)}}} & \rotatebox{90}{\makecell[l]{\textbf{fence} \\ \scriptsize{(3.09\%)}}} & \rotatebox{90}{\makecell[l]{\textbf{pole} \\ \scriptsize{(0.29\%)}}} 
& \rotatebox{90}{\makecell[l]{\textbf{traffic sign} \\ \scriptsize{(0.08\%)}}} \\
& & 
& \textcolor[RGB]{128,0,128}{\rule{6pt}{6pt}} & \textcolor[RGB]{255,0,255}{\rule{6pt}{6pt}} & \textcolor[RGB]{255,182,193}{\rule{6pt}{6pt}}
& \textcolor[RGB]{139,0,0}{\rule{6pt}{6pt}} & \textcolor[RGB]{255,215,0}{\rule{6pt}{6pt}} & \textcolor[RGB]{0,0,255}{\rule{6pt}{6pt}}
& \textcolor[RGB]{0,191,255}{\rule{6pt}{6pt}} & \textcolor[RGB]{173,216,230}{\rule{6pt}{6pt}} & \textcolor[RGB]{199,21,133}{\rule{6pt}{6pt}}
& \textcolor[RGB]{65,105,225}{\rule{6pt}{6pt}} & \textcolor[RGB]{34,139,34}{\rule{6pt}{6pt}} & \textcolor[RGB]{139,69,19}{\rule{6pt}{6pt}}
& \textcolor[RGB]{144,238,144}{\rule{6pt}{6pt}} & \textcolor[RGB]{255,69,0}{\rule{6pt}{6pt}} & \textcolor[RGB]{219,112,147}{\rule{6pt}{6pt}}
& \textcolor[RGB]{75,0,130}{\rule{6pt}{6pt}} & \textcolor[RGB]{112,128,144}{\rule{6pt}{6pt}} & \textcolor[RGB]{218,165,32}{\rule{6pt}{6pt}}
& \textcolor[RGB]{255,228,181}{\rule{6pt}{6pt}} \\
\midrule
MonoScene~\cite{cao2022monoscene} & C & 10.9 & 54.2 & 26.4 & 24.1 & 4.8 & 14.2 & 18.7 & 3.6 & 0.7 & 0.9 & 4.2 & 15.3 & 2.7 & 19.2 & 1.6 & 0.2 & 0.3 & 11.5 & 3.1 & 2.0 \\
SurroundOcc~\cite{wei2023surroundocc} & C & 11.7 & 56.1 & 27.9 & 31.1 & 6.3 & 14.8 & 21.0 & 1.2 & 1.3 & 1.1 & 4.6 & 15.1 & 4.3 & 19.1 & 1.5 & 2.2 & 0.2 & 10.1 & 4.3 & 1.0 \\
OccFormer~\cite{zhang2023occformer} & C & 12.9 & 55.7 & 29.3 & 30.5 & 15.5 & 21.0 & 22.2 & 1.1 & 1.2 & 1.4 & 4.6 & 15.3 & 4.3 & 19.2 & 1.3 & 2.5 & 0.6 & 13.2 & 4.1 & 1.8 \\
RenderOcc~\cite{pan2024renderocc} & C & 13.1 & 57.0 & 29.2 & 32.4 & 16.6 & 19.6 & 24.8 & 6.4 & 2.7 & 0.2 & 3.6 & 26.2 & 5.8 & 3.6 & 0.0 & 1.1 & 0.6 & 9.3 & 6.1 & 3.4 \\
LMSCNet~\cite{roldao2020lmscnet} & L & 14.3 & 64.1 & 34.2 & 29.4 & 6.3 & 39.2 & 27.4 & 2.2 & 0.0 & 0.3 & 2.2 & 19.3 & 3.2 & 19.4 & 0.3 & 0.0 & 0.1 & 15.3 & 5.1 & 3.5 \\
JS3C-Net~\cite{yan2021sparse} & L & 22.8& 64.2 & 38.4 & 34.6 & \textbf{13.3} & 39.5 & 34.6 & 7.9 & 15.2 & 8.0 & 12.2 & 42.3 & 18.2 & 39.3 & 6.1 & 5.1 & 0.5 & 30.7 & 17.3 & 5.2 \\
SSC-RS~\cite{mei2023ssc} & L & 24.2 & 71.4 & 43.8 & 41.2 & 11.2 & \textbf{44.3} & 36.1 & 5.4 & 13.2 & 4.5 & \textbf{14.2} & \textbf{43.5} & 25.2 & \textbf{43.1} & 2.4 & 1.5 & 0.4 & 36.7 & 16.2 & 6.1 \\
M-CONet~\cite{wang2023openoccupancy} & C\&L & 20.5 & 61.6 & 38.3 & 29.2 & 13.2 & 38.3 & 33.5 & 4.3 & 3.1 & 2.4 & 5.5 & 41.7 & 20.2 & 35.4 & 0.7 & 2.2 & 0.4 & 26.1 & 18.3 & 15.6 \\
Co-Occ~\cite{pan2024co} & C\&L & 23.9 & 71.6 & 41.8 & \textbf{42.2} & 10.3 & 35.4 & 39.8 & 5.7 & 4.2 & 3.2 & 7.9 & 41.1 & 30.4 & 39.8 & 1.2 & 3.1 & 0.3 & 31.4 & \textbf{25.7} & 19.7 \\
\midrule
\textbf{TACOcc (Ours)} & C\&L & \textbf{24.7} & \textbf{72.4} & \textbf{44.1} & 34.5 & 8.2 & 32.1 & \textbf{40.9} & \textbf{8.6} & \textbf{15.2} & \textbf{8.3} & 8.8 & 34.0 & \textbf{30.8} & 40.9 & \textbf{6.3} & \textbf{5.4} & \textbf{0.8} & \textbf{36.9} & 20.4 & \textbf{20.7} \\
\bottomrule
\end{tabular}
}
}
\end{table}

To further validate the effectiveness of our framework, we conduct a comparative analysis with SOTA methods~\cite{cao2022monoscene,mei2023ssc,pan2024co,pan2024renderocc,roldao2020lmscnet,wang2023openoccupancy,wei2023surroundocc,yan2021sparse,zhang2023occformer}on the SemanticKITTI test set ~\cite{behley2019semantickitti} (Tab.\ref{tab:result2}). As shown in the table, our method outperforms JS3CNet ~\cite{yan2021sparse} by 1.9\% mIoU and SSC-RS ~\cite{mei2023ssc} by 0.5\% mIoU, despite their use of additional lidar segmentation supervision.

\subsection{Ablation Studies}

The TACOcc framework systematically validates the synergistic optimization mechanism of dynamic multi-modal fusion and volume rendering supervision in 3D semantic occupancy prediction through ablation experiments. Experimental results (Tab.\ref{tab:table3}) demonstrate that the baseline model achieves 34.3\% IoU and 22.1\% mIoU on the nuScenes dataset. When employing fixed neighborhood sampling fusion, significant geometric-semantic feature misalignment emerges for small-scale objects (e.g., pedestrians). After introducing the target scale-adaptive bidirectional symmetric retrieval mechanism, the dynamic neighborhood ranges prediction driven by queries elevates performance to 38.7\% IoU (+4.4\%) and 26.9\% mIoU (+4.8\%), substantially improving multi-modal alignment accuracy and validating the mechanism's effectiveness in mitigating feature mismatch. Furthermore, the differentiable pipeline rigorously maps 3D feature fields to multi-view 2D spaces through projective transformation. By jointly optimizing the photometric reprojection loss \( \mathcal{L}_{\text{rgb}} \)  and Gaussian parameter consistency loss \( \mathcal{L}_{\text{pc}} \), a 2D-3D supervision is established. The model achieves SOTA performance with an IoU of 41.8\% and an mIoU of 28.4\%. It confirms that this dual supervision strategy balances robustness perception and fine-grained reconstruction enhancement in complex scenarios.

\begin{table}[h]
    \centering
    \vspace{-0.5em}
    \begin{minipage}{0.45\textwidth}
        \centering
        \small
        \captionsetup{justification=centering}
        \caption{The impact of different components.}
        \label{tab:table3}
        \vspace{0.5em}
        {\setlength{\tabcolsep}{3pt}
        \begin{tabular}{@{}c c c c c c c@{}}
            \toprule
            \multirow{2}{*}{Base} & \multicolumn{2}{c}{Fusion} & \multicolumn{2}{c}{Rendering} & \multirow{2}{*}{IoU} & \multirow{2}{*}{mIoU} \\ \cmidrule(lr){2-3} \cmidrule(lr){4-5}
            & Fix $k$ & Dynamic $k$ & \(\mathcal{L}_{\text{rgb}}\) & \(\mathcal{L}_{\text{pc}}\) & & \\ \midrule
            \(\checkmark\) & & & & & 34.3 & 22.1 \\
            \(\checkmark\) & \(\checkmark\) & & & & 36.6 & 25.3 \\
            \(\checkmark\) & & \(\checkmark\) & & & 38.7 & 26.9 \\
            \(\checkmark\) & & \(\checkmark\) & \(\checkmark\) & & 40.2 & 27.1 \\
            \(\checkmark\) & & \(\checkmark\) & \(\checkmark\) & \(\checkmark\) & \textbf{41.8} &  \textbf{28.4} \\ \bottomrule
        \end{tabular}
        }
    \end{minipage}
    \hfill
    \begin{minipage}{0.45\textwidth}
        \centering
        \small
        \captionsetup{justification=centering}
        \caption{Fusion parameter selection.}
        \label{tab:table4}
        \vspace{0.5em}
        \begin{tabular}{@{}c c c c@{}}
            \toprule
            $k$ & IoU$\uparrow$ & mIoU$\uparrow$ & Latency(s)\\
            \midrule
            1 & 35.4 & 24.9 & 0.52 \\
            2 & 37.1 & 25.6 & 0.59\\
            3 & 38.2 & 26.7 & 0.65\\
            1 - 2 & 37.8 & 25.4 & 0.55\\
            1 - 3 & 39.4 & 27.2 & 0.60 \\
            1 - 4 & \textbf{41.8} & \textbf{28.4} & 0.64\\
            \bottomrule
        \end{tabular}
    \end{minipage}
\vspace{-0.5em}
\end{table}

We evaluate the impact of the neighborhood retrieval range (Tab.\ref{tab:table4}). Experimental results show that the fixed $k$ strategy improves accuracy but reduces efficiency rapidly. However, the dynamic range strategy achieves a balance between accuracy and efficiency. When $k$ is set to 1-4, the IoU and mIoU reach 41.8\% and 28.4\%, respectively. This represents an improvement of 3.6\% and 1.7\% compared to fixed $k$ = 3, while the latency remains low at only 0.64s on a single RTX A6000 GPU.

\begin{table}[h]
    \centering
    \begin{minipage}{0.45\textwidth}
    \vspace{-1em}
        \centering
        \captionsetup{justification=centering}
        \caption{Frequencies of different $k$ with more small targets (136,589 non-zero queries).}
        \label{table5}
        \vspace{0.5em}
        \begin{tabular}{ccccc}
            \toprule
            $k$ & 1 & 2 & 3 & 4 \\
            \midrule
            Freq. & 38261 & 43814 & 32378 & 22136 \\
            Prob. & \textbf{28.0\%} & 32.1\% & 23.7\% & 16.2\% \\
            \bottomrule
        \end{tabular}
    \end{minipage}
    \hfill
    \begin{minipage}{0.45\textwidth}
    \vspace{-1em}
        \centering
        \captionsetup{justification=centering}
        \caption{Frequencies of different $k$ with more large targets (129,548 non-zero queries).}
        \label{table6}
        \vspace{0.5em}
        \begin{tabular}{ccccc}
            \toprule
            $k$ & 1 & 2 & 3 & 4 \\
            \midrule
            Freq. & 14539 & 23953 & 40432 & 50624 \\
            Prob. & 11.2\% & 18.5\% & 31.2\% & \textbf{39.1\%} \\
            \bottomrule
        \end{tabular}
    \end{minipage}
\end{table}

\begin{figure}[h]
    \centering
    \begin{minipage}{0.45\textwidth}
    \vspace{-0.5em}
        \centering
        \includegraphics[
            width=0.6\textwidth,
            trim={0 0 0 0},
            clip
        ]{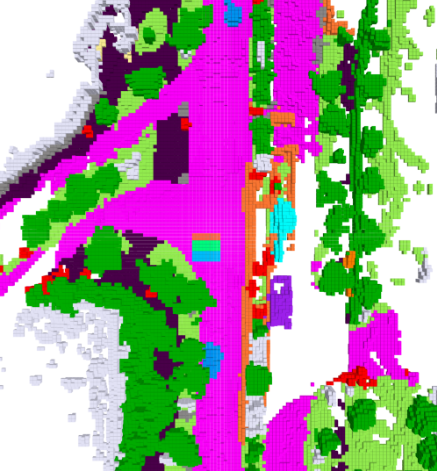}
        \caption{Scenarios with more small targets.}
        \label{fig:figure4}
    \end{minipage}
    \hfill
    \begin{minipage}{0.45\textwidth}
    \vspace{-0.5em}
        \centering
        \includegraphics[
            width=0.6\textwidth,
            trim={0 0 0 0},
            clip
        ]{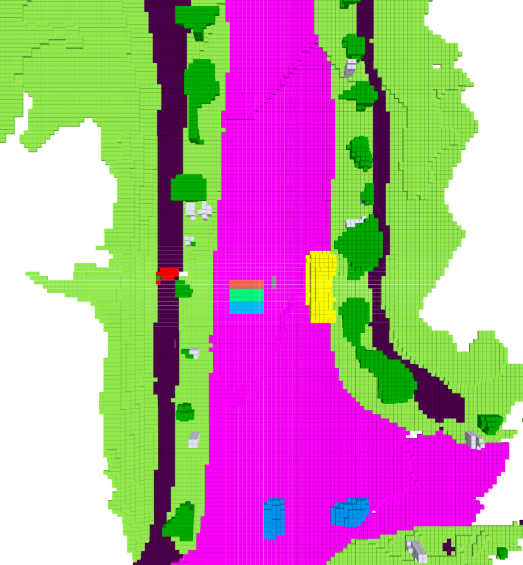}
        \caption{Scenarios with more large targets.}
        \label{fig:figure5}
    \end{minipage}
\end{figure}

A comparison of the distribution of $k$ in different target scenarios shows that in scenarios (Tab.~\ref{table5},~\ref{table6}, Fig.~\ref{fig:figure4},~\ref{fig:figure5}) dominated by small targets, the proportion of $k$ = 1-2 reaches 60.1\%, indicating that small targets tend to use a narrow neighborhood to suppress noise. In contrast, in scenarios with large targets, the proportion of $k$ = 3-4 accounts for 70.3\% while that of $k$ = 1 is only 11.2\%, suggesting that large targets require an expanded neighborhood to enhance context understanding. This difference validates the effectiveness of the scale-perception mechanism in the dynamic strategy.

\section{Conclusion and Limitations}
We propose TACOcc, it addresses challenges in 3D semantic occupancy prediction through a novel adaptive multi-modal fusion framework. First, it incorporates a bidirectional feature alignment module that dynamically adjusts neighborhood ranges to enhance multi-modal matching accuracy. Second, a volume rendering optimization module leverages image photometric constraints to refine 3D Gaussian parameters, improving detail reconstruction. Evaluated on the nuScenes dataset, TACOcc achieves a SOTA 28.4\% mIoU, offering an efficient and scalable solution for autonomous driving perception.This study has the following limitations: (1) The dynamic feature fusion module may have limited perception of extreme-sized objects, which can be addressed by adjusting the range of candidate $k$ to successfully perceive objects of extreme sizes. (2) The volume rendering module depends on precise sensor calibration and is vulnerable to environmental factors, which may lead to accuracy degradation.

{\small
\bibliographystyle{plainnat}
\bibliography{refs}
}


\appendix

\section{Technical Appendices and Supplementary Material}

We compares the performance of different methods in 3D semantic occupancy prediction at distances from 25m to 100m (Tab.\ref{table8}). TACOcc outperforms other methods in both short and long range scenarios, demonstrating strong performance. The results confirm the effectiveness of its dynamic approach in complex environments.

\begin{table}[h]
\centering
\caption{3D Semantic occupancy results with different ranges.}
\label{table8}
\begin{tabular}{lccccccc}
\toprule
\multirow{2}{*}{Method} & \multicolumn{3}{c}{IoU} & \multicolumn{3}{c}{mIoU} \\
\cmidrule(lr){2-4} \cmidrule(lr){5-7}
 & 25m & 50m & 100m & 25m & 50m & 100m\\
\midrule
M-CONet~\cite{wang2023openoccupancy} & 60.7 & 51.1 & 29.0 & 36.7 & 31.4 & 23.6 \\
Co-Occ~\cite{pan2024co} & 62.3 & 52.9 & 41.0 & 39.3 & 34.2 & 26.6 \\
TACOcc(ours) & 63.8 & 54.9 & 43.2 & 41.8 & 36.5 & 28.4 \\
\midrule
Improvements (\%) & \textbf{+1.5} & \textbf{+2.0} & \textbf{+2.2} & \textbf{+2.5} & \textbf{+2.3} & \textbf{+1.8} \\
\bottomrule
\end{tabular}
\end{table}

We evaluated the inference time and memory usage of different image sizes and 2D backbone networks on a single RTX A6000 GPU (Tab.\ref{table7}). For the model with low image resolution and ResNet50, its computational cost and latency are low, but its performance is not high. Increasing the image size and deepening the ResNet do not significantly increase the memory usage and latency, and can greatly improve the performance.  
\begin{table}[h]
\centering
\caption{Efficiency analysis.}
\label{table7}
\begin{tabular}{lcccccc}
\toprule
Image Size & 2D backbone & IoU & mIoU & Memory (G) & Latency (s) \\
\midrule
256 × 704 & R50 & 38.9 & 24.8 & 10.78 & 0.47 \\
896 × 1600 & R50 & 40.3 & 27.2 & 11.69 & 0.55 \\
896 × 1600 & R101 & \textbf{41.8} & \textbf{28.4} & 11.78 & 0.64 \\
\bottomrule
\end{tabular}
\end{table}

\begin{figure}[h]
\includegraphics[
    width=1\textwidth,
    trim={120pt 50pt 100pt 50pt}, 
    clip
]{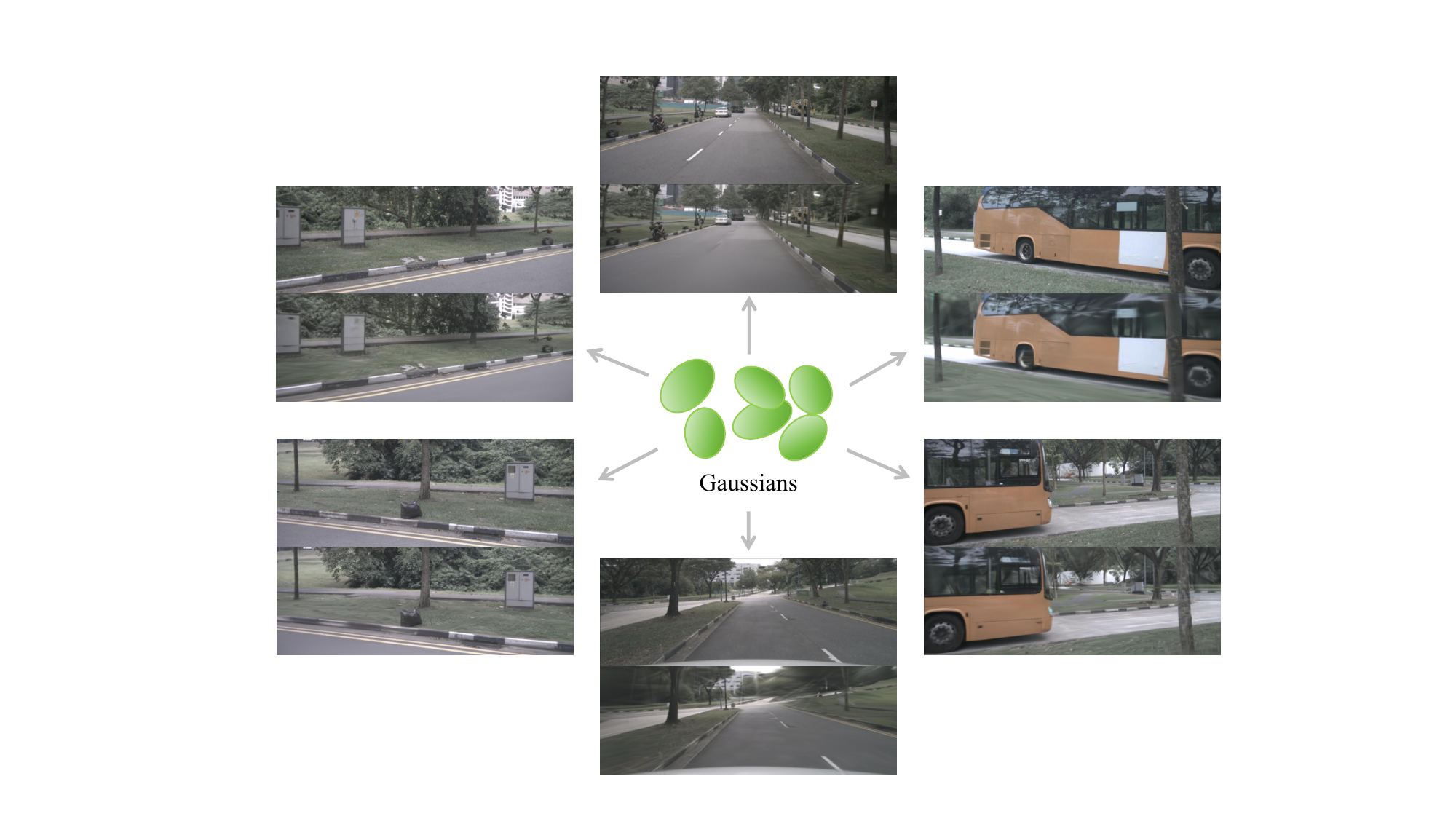}
\centering
\caption{Rendered images display and comparison with real images. For each viewpoint, the image above is the real one, and the image below is the rendered result.}
\label{fig:figurerender}
\vspace{0.3em}
\end{figure}

We present a comparison of rendering results from the volume rendering optimization module. As shown in Fig.\ref{fig:figurerender}, the upper part displays the input images of real scenes, while the lower part presents the volume-rendered outputs from corresponding viewpoints. From a visual perspective, the rendered results demonstrate high fidelity in geometric detail reconstruction, material surface reflection properties, and consistency under complex lighting conditions.

\begin{figure}[h]
\includegraphics[
    width=1\textwidth,
    trim={240pt 0 240pt 0}, 
    clip
]{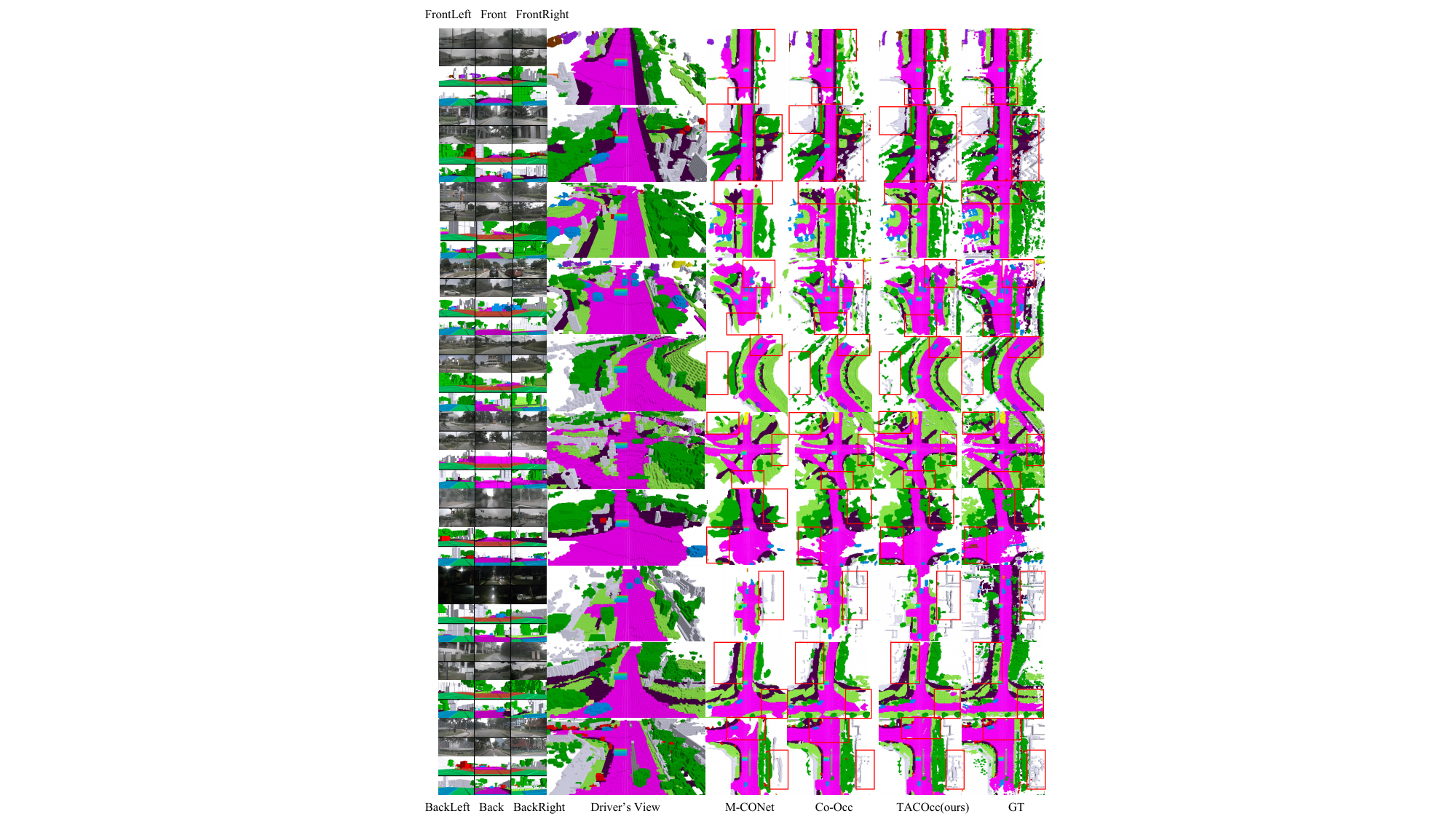}
\centering
\begin{scriptsize}
\textcolor[RGB]{255,0,255}{\rule{5pt}{5pt}} driveable surface \quad
\textcolor[RGB]{0,150,245}{\rule{5pt}{5pt}} car \quad
\textcolor[RGB]{255,255,0}{\rule{5pt}{5pt}} bus  \quad
\textcolor[RGB]{160,32,240}{\rule{5pt}{5pt}} truck \quad
\textcolor[RGB]{150,240,80}{\rule{5pt}{5pt}} terrain \quad
\textcolor[RGB]{0,175,0}{\rule{5pt}{5pt}} vegetation \quad
\textcolor[RGB]{75,0,75}{\rule{5pt}{5pt}} sidewalk \quad
\textcolor[RGB]{139,137,137}{\rule{5pt}{5pt}} other flat \quad
\textcolor[RGB]{255,0,0}{\rule{5pt}{5pt}} pedestrian \quad
\textcolor[RGB]{255,192,203}{\rule{5pt}{5pt}} bicycle \quad
\textcolor[RGB]{230,230,250}{\rule{5pt}{5pt}} manmade \quad
\textcolor[RGB]{200,180,0}{\rule{5pt}{5pt}} motorcycle \quad
\textcolor[RGB]{255,120,50}{\rule{5pt}{5pt}} barrier \quad
\textcolor[RGB]{0,255,255}{\rule{5pt}{5pt}} construction vehicle \quad
\textcolor[RGB]{135,60,0}{\rule{5pt}{5pt}} trailer \quad
\textcolor[RGB]{255,240,150}{\rule{5pt}{5pt}} traffic cone \quad
\vbox{%
  \offinterlineskip
  \hbox{\color[RGB]{227,83,60}\vrule width 5pt height 1.66666pt depth 0pt}
  \hbox{\color[RGB]{0,220,110}\vrule width 5pt height 1.66667pt depth 0pt}
  \hbox{\color[RGB]{0,166,221}\vrule width 5pt height 1.66667pt depth 0pt}
} ego vehicle
\end{scriptsize}
\caption{Additional qualitative comparison results on the nuScenes validation set. The top left displays the input surrounding images, while the bottom left shows the occupancy predictions made by our TACOcc for each surrounding image. Subsequently visualized are the driver's perspective from our TACOcc method, the BEV of M-CONet~\cite{wang2023openoccupancy}, the BEV of Co-Occ~\cite{pan2024co}, the BEV of our TACOcc method, and the ground truth (GT) occupancy from SurroundOcc~\cite{wei2023surroundocc}.}
\label{figure6}
\end{figure}

Due to space limitations, additional visualization results on the nuScenes dataset are provided in Fig.~\ref{figure6}. The comparisons with two representative multi-modal methods highlight the effectiveness of our proposed TACOcc.

\end{document}